\documentclass[11pt,a4paper]{article}
\usepackage[hyperref]{acl2019}
\usepackage{times}
\usepackage{latexsym}

\usepackage{url}
\usepackage{graphicx}
\aclfinalcopy 

\begin{document}
\title{Reporting the Unreported: Event Extraction for Analyzing the Local Representation of Hate Crimes}
%
\author{\\{\bf Aida Mostafazadeh Davani,
    Leigh Yeh,
    Mohammad Atari,
    Brendan Kennedy} \\
    {\bf Gwenyth Portillo-Wightman,
    Elaine Gonzalez,
    Natalie Delong,
    Rhea Bhatia,} \\
    {\bf Arineh Mirinjian, 
    Xiang Ren,
    Morteza Dehghani} \AND
    {\normalfont University of Southern California}
 }
\date{}

\maketitle
\begin{abstract}
 Official reports of hate crimes in the US are under-reported relative to the actual number of such incidents. Further, despite statistical approximations, there are no official reports from a large number of US cities regarding incidents of hate. Here, we first demonstrate that event extraction and multi-instance learning, applied to a corpus of local news articles, can be used to predict instances of hate crime. We then use the trained model to detect incidents of hate in cities for which the FBI lacks statistics. 
 Lastly, we train models on predicting homicide and kidnapping, compare the predictions to FBI reports, and establish that incidents of hate are indeed under-reported, compared to other types of crimes, in local press.  
\end{abstract}

\section{Introduction}

Hate crimes are defined as crimes of violence either against a person or their property that display evidence of prejudice based on the victims' race, gender or gender identity, religion, disability, sexual orientation, or ethnicity \cite{jacobs1998hate}. According to the results of a new Department of Justice hate crime report released in 2017 \cite{masucci2017hate}, approximately 54\% of hate crime victimizations were not reported to police during 2011-2015. Despite the recent efforts of advocacy groups, policy makers, and researchers to create reliable, national data to understand the extent and severity of hate crime victimization, the existing estimates continue to fall short \citep{pezzella2019dark}.

It stands to reason that hate crimes induce local disturbance, and as a result, might be likely to get local coverage. Therefore, local news agencies can be considered a unique source of information for detecting these incidents. Here, we use a corpus of local news articles, collected from the Patch\footnote{https://www.patch.com} website. The Patch data contain independent, hyper-local news articles compiled from local news sites.

We apply event extraction methods to identify incidents of hate crime reported in the Patch corpus for cities with no representation in FBI reports, and analyze the frequency of the extracted events compared to the number of incidents reported by the FBI. The task of labeling each article as a hate crime or not is defined as a Multi-Instance Learning (MIL) problem since each article is modeled as a sequence of sentences. Instead of predicting a label for each sentence, we use the information embedded in all the sentences of an article to determine whether the article is reporting a hate crime. 

After testing the model on a set of annotated articles, we apply the trained model to cities for which the FBI does not have any reports, and we provide a lower-bound estimate on the occurrence frequency of hate crimes in those cities. Lastly, we compare the coverage of incidents of hate as reported in local news sources with coverages of two non-hate crimes, namely homicides and kidnappings, and contrast the overlap of the extracted incidents with those reports by the FBI.

Our results show that applying MIL for event extraction can help approximate the missing reports, especially in cases in which publishing the comprehensive event set faces challenges and is influenced by subjective bias.



\section{MIL for Event Extraction} \label{sec:method}

\begin{figure*}[t]
    \centering
    \includegraphics[scale=0.5]{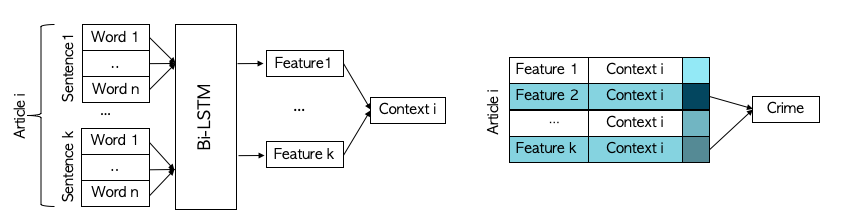}
    \caption{The event detection model using a MIL network. Local representation of each sentence are combined with context representation of its related article.}
    \label{fig:detect}
\end{figure*}

In this paper, we perform event detection and extraction on news articles based on taxonomies of acts of crime. We adapt the MIL approach for event detection developed by \citet{wang2016multiple}, which identifies key sentences for a given article. We then use these key sentences to perform event extraction, predicting the target and type of action for a given incident.

\smallskip
\noindent
\textbf{Event Detection}


The MIL approach for document classification is illustrated in Figure \ref{fig:detect}. The two basic components are the creation of local features (representations of sentences) and the aggregation of these features into a document representation. 
Whereas \citet{wang2016multiple} applies Convolutional Neural Networks (CNNs) for the creation of local features, we use a bidirectional Long Short-Term Memory \citep[LSTM;][]{hochreiter1997long} network for representing each sentence of an article. Bidirectional networks \citep{graves2005framewise} have been shown to provide a good semantic representation of textual data \citep{huang2015bidirectional}. 

Local representations are then aggregated to form a ``contextual'' representation of the document, using a CNN layer. This context vector, which is the same for all sentences in the document, is then concatenated with \emph{each} sentence's local representation.

Given the feature representation of the sentences in an article, the probabilistic score of each sentence in an article is calculated using a fully connected layer with sigmoid activation. This probabilistic score shows the extent to which the sentence contributes to predicting the crime label of the article.
The label for a bag of sentences is calculated by averaging the \textit{k} highest probabilistic scores. We checked the results with \textit{k} being set to 2 or 3, since a few number of sentences in each article can determine the label.

Another prominent method which we compared the MIL results with is Hierarchical Attention Networks \citep[HAN;][]{yang2016hierarchical}. HANs apply attention first at the level of words, then at the level of sentences, to produce representations of documents subject to local variations in textual importance. We also compare the results of neural network models with TF-IDF as a text classification baseline.


\smallskip
\noindent
\textbf{Event Extraction}

The most challenging aspect of extracting events from a sentence is that the context of a document should be considered in order to interpret an entity and the type of triggered event \citep{chen2015event}. Approaches that exclusively use word features for the task usually lack comprehensiveness.

The event detection model in the previous section produces, for each positive prediction, a small set of sentences likely to influence the document's label. In the event extraction step, we use a bidirectional LSTM text classifier to predict the attributes of a crime event.

The attributes of a crime event are determined by the taxonomy proposed by \citet{kennedy2018typology} for annotating hate rhetoric. In our case (see Section \ref{data}), we are predicting two attributes: the target of a crime event, and the type of crime. 

Formulated as a multi-class, multi-task prediction, we train a biLSTM to produce a representation of the concatenation of the top two sentences and feed this to two separate feed-forward networks, one predicting the target categorization and one the crime type.


    

\section{Data} \label{data}
The Patch website includes hyper-local news articles from 1217 cities based in the US. For this project, we scraped the articles in the ``Fire and Crime'' category of Patch, resulting in a corpus containing $\sim370k$ unlabeled local news articles. For our experiments, we manually annotate subsets of the main dataset for training event detection models.


Our annotations consisted of a binary label --- whether the article represents a specific hate crime --- as well as labeling the attributes of hate crime articles, which consist of the target of the action (whether the crime was based on the race, nationality, gender, religion, sexual orientation, ideology, political identification or mental/physical health of the target) and the type of action (whether the crime was an assault, arson, vandalism or hate demonstration).

For gathering a subset of articles for annotation, we filtered the news articles based on a set of 8 keywords (\textit{swastika}, \textit{hate}, \textit{racial}, \textit{religion}, \textit{religious}, \textit{gay}, \textit{transgender}, \textit{transsexual}) related to hate crimes, resulting in $\sim3k$ patch articles, which were then combined with $500$ randomly sampled articles to account for the high frequency of the hate crimes in the selected dataset. Each article was annotated for the presence and the attributes of the hate crime reports by one annotator. Annotators achieved $0.73$ inter-coder agreement on a subset of $500$ posts based on Cohen's Kappa \citep{cohen1968weighted}.

For hate crime articles that are not associated with the keywords, we expected the model's predictions to be sparse. To deal with this problem we applied an active learning approach introduced by \citet{lewis1994sequential}. In this approach, after training the model, we predicted the hate crime label for all the articles in the dataset and gathered their associated probabilities. 
We then selected $\sim1k$ articles based on their probability score, using a normal distribution with a mean of 0.5 and standard deviation of 0.1. This set of articles, for which the model was uncertain about their labels, was then annotated by the same annotators and added to the training set.  

We performed a similar procedure, without entity labeling and active learning, for homicide (keywords: \textit{homicide}, \textit{manslaughter}, \textit{murder}, and \textit{kill}) and kidnapping (keywords: \textit{kidnapping}, \textit{abduct}, \textit{hostage}, \textit{abduct}, and \textit{shanghai}) events. The frequency statistics for these annotations are represented in Table \ref{tab:datastats}.

\begin{table}[h]
    \centering
    \begin{tabular}{c|c|c}
         Event Type & Positive & Negative \\
         \hline
         Hate Crime & 1979 & 3192 \\
         Homicide & 1664 & 1327 \\
         Kidnapping & 1864 & 1104 \\ 
    \end{tabular}
    \caption{Frequency of events from Patch annotations. 
    }
    \label{tab:datastats}
\end{table}

Type and target of the hate crime was also annotated for each article. Crime type labels are distributed across assaults ($900$), arson ($76$), vandalism ($450$), and hate demonstrations ($543$). The most frequent target types were race ($1029$), religion ($376$), and sexual orientation ($265$).

\section{Experiment}
\label{sec:experiment}

All models were implemented with Tensorflow \citep{abadi2016tensorflow}. Hidden size of the LSTM cells was set to 50, filter sizes of the CNN were set to 2, 3 and 4, and a dropout layer was placed on top of the LSTM cell to set 25\% of the values to zero. Each batch included 5 articles converted to their latent representation using 300-dimensional GloVe word embeddings \citep{pennington2014glove}. Parameter tuning was performed with 70\% of the dataset as the train set and 10\% as development set and the learning rate was set to 0.00008.

All three models for predicting hate crime, kidnapping and homicide were trained for 50 epochs. 

\section{Results}

\begin{table}[t]
    \centering
    \begin{tabular}{c|c|c|c}
         & MIL & HAN & TF-IDF \\ \hline
        Hate crime & \textbf{82.9} & 82.6 & 81.6 \\
        Homicide & \textbf{81.3} & 79.7 & 77.4 \\
        Kidnapping & \textbf{78.7} & 75.6 & 73.9 \\
    \end{tabular}
    \caption{Event detection F1 scores for the test set}
    \label{tab:accuracy}
\end{table}

The resulting F1-scores are calculated for the test set and represented in Table \ref{tab:accuracy}.

We apply the learned models to make predictions about the rate of hate crimes in cities for which the FBI lacks data. We also compare the relative rate of news coverage of hate crime with those of homicides and kidnappings. 

\smallskip
\noindent
\textbf{Predicting Hate Crime} 

First, we compare the positive hate crime labels predicted for Patch with the FBI's city-level hate crime reports. 
After applying the trained model to the Patch dataset, we captured 3152 articles that report hate crime incidents. These articles include 678 reports from 286 cities that have no representation in the FBI reports. This suggests that the MIL model applied to the local news dataset can approximate missing statistics on hate crime in those cities. However, presuming a one-to-one relation between the news articles and hate crime incidents is not accurate, since there can be false positive results and duplicated articles about an incident. To provide an accurate set of unreported hate crime incidents we removed duplicated and misclassified articles from the set of 678 unrepresented hate crime incidents.

In order to account for the possible duplications, we utilize the event extraction model to capture the event entities, namely target and action type. Running the extraction model with the same hyperparameters yields the results presented in Table \ref{tab:accuracy_extraction}. We use the entities together with the time (mentioned in the dataset) and location (extracted with named entity recognizer of CoreNLP \citep{manning2014stanford}) of the articles to detect duplicated events.
\begin{table}[h]
    \centering
    \begin{tabular}{c|c|c|c}
        Label & Precision & Recall & F1 \\ \hline
        Target & 63.9 & 65.3 & 63.9 \\
        Action & 67.7 & 68.0 & 67.4
    \end{tabular}
    \caption{Event extraction scores of MIL}
    \label{tab:accuracy_extraction}
\end{table}

After checking for pairs of articles from the same state and city, with the same reported target victim and crime action, reported at most one day apart from each other, we found 20 pairs of duplicated articles, indicating 658 unique incidents of hate in the cities with no representation in the FBI dataset.

Next, we manually checked these articles and found 315 articles that were correctly labeled as hate crime. Table \ref{tab:false_pos} represents a few instances of false positives. Exploring the false positive results indicates that non-hate crime articles that mention minority social groups are often incorrectly labeled as hate crime.  This issue can be explored further in future works to improve the accuracy of the predictions.

\begin{table}[]
    \centering
    \begin{tabular}{c|p{6cm}}
         & False Positive Examples \\ \hline
        1 & A former Ku Klux Klan leader in Ozark was sentenced Thursday to a decade in prison for sexually abusing a woman in southern Alabama. \\
        2 & The FBI is part of an investigation into a suspicious substance delivered to a Council on American-Islamic Relations office in Santa Clara on Thursday.\\
        3 & The hate from the violent white nationalist gathering that resulted in the death of an anti-racism protester in Charlottesville, can be found anywhere. \\
    \end{tabular}
    \caption{First sentences of sample articles recognized as false positive results by our annotators.}
    \label{tab:false_pos}
\end{table}

\smallskip
\noindent
\textbf{Comparisons to Other Crime}

In order to compare the coverage of incidents of hate with coverage of homicides and kidnappings, we contrast the overlap of the extracted incidents with those reported by the FBI. Specifically, for 159 cities that have representation of the three crimes both in Patch and FBI crime reports, we calculate the ratio of Patch-based predictions to FBI reports for each crime.

To investigate the differences between the distributions of these ratios, we ran a Welch-type one-way ANOVA, which is robust to non-normal distributions, allowing for heteroscedasticity and extreme non-normality of ratios in our data \citep{field2017robust}. The results indicates that the three crimes' distributions have significantly different medians (\textit{F}[2,214.28] = 102.03, \textit{p} $\textless$ 0.001). Post-hoc tests suggested that Patch-based estimates of hate are significantly lower than homicide and kidnapping (both \textit{p}'s $\textless$ 0.001).

\section{Discussion}
Hate crimes in the US remain vastly under-reported \cite{masucci2017hate}. For instance, only 12.6\% of the agencies in the FBI report indicated that hate crimes had occurred in their jurisdictions in 2017, and agencies as large as the Miami Police Department reported zero incidents of hate \citep{FBIreport2017}, which seem unrealistic. The contributions of this paper are two-fold: First, we have shown that event detection can be applied to the study of hate crimes. 
Specifically, we demonstrated that using MIL for event detection, in conjunction with local news articles, can provide conservative estimates of the occurrence of hate crimes in cities with no official representation. A possible application of this method is creating a real-time hate crime detector based on online local news agencies, providing researchers and community workers a lower-bound on the number of hate crimes in such locations.  Second, the statistical analyses suggested that the scope of hate crime coverage in local news is lower than that of violent, but non-hate crimes. This suggests that although local news sources can be used as an additional source for gathering better statistics about hate crimes, the predictions of our models are simply lower bound estimates.




\bibliographystyle{acl_natbib.bst}
\bibliography{acl2018}
\end{document}